\documentclass[letterpaper]{article} 
\usepackage{aaai2026}  
\usepackage{times}  
\usepackage{helvet}  
\usepackage{courier}  
\usepackage[hyphens]{url}  
\usepackage{graphicx} 
\usepackage{booktabs}  
\usepackage{multirow}  
\usepackage{colortbl}  
\usepackage[table,xcdraw]{xcolor} 
\usepackage{booktabs}
\usepackage[table,xcdraw]{xcolor}
\usepackage{adjustbox}
\usepackage{amsmath}
\usepackage{amsfonts}
\usepackage{booktabs}
\usepackage[table]{xcolor}
\usepackage{graphicx} 
\usepackage{pifont}
\usepackage{makecell}

\definecolor{first}{HTML}{C0392B}  
\definecolor{second}{HTML}{2471A3}  
\usepackage{natbib}  
\usepackage{caption} 
\usepackage{algorithm}
\usepackage{algorithmic}

\usepackage{newfloat}
\usepackage{listings}
\DeclareCaptionStyle{ruled}{labelfont=normalfont,labelsep=colon,strut=off} 
\floatstyle{ruled}
\newfloat{listing}{tb}{lst}{}
\floatname{listing}{Listing}
\nocopyright

\title{UniLDiff: Unlocking the Power of Diffusion Priors for All-in-One Image Restoration}

\author {
    Zihan Cheng\textsuperscript{\rm 1},
    Liangtai Zhou\textsuperscript{\rm 1},
    Dian Chen\textsuperscript{\rm 1},
    Ni Tang\textsuperscript{\rm 1},
    Xiaotong Luo\textsuperscript{\rm 1},
    Yanyun Qu\textsuperscript{\rm 1}
}
\affiliations {
    \textsuperscript{\rm 1}School of Informatics, Xiamen University\\

}

\usepackage{bibentry}

\begin{document}

\maketitle

\begin{abstract}

All-in-One Image Restoration (AiOIR) has emerged as a promising yet challenging research direction. To address the core challenges of diverse degradation modeling and detail preservation, we propose UniLDiff, a unified framework enhanced with degradation- and detail-aware mechanisms, unlocking the power of diffusion priors for robust image restoration. Specifically, we introduce a Degradation-Aware Feature Fusion (DAFF) to dynamically inject low-quality features into each denoising step via decoupled fusion and adaptive modulation, enabling implicit modeling of diverse and compound degradations. Furthermore, we design a Detail-Aware Expert Module (DAEM) in the decoder to enhance texture and fine-structure recovery through expert routing. Extensive experiments across multi-task and mixed degradation settings demonstrate that our method consistently achieves state-of-the-art performance, highlighting the practical potential of diffusion priors for unified image restoration. Our code will be released.

\end{abstract}


\section{Introduction}

Traditional image restoration (IR) focuses on a specific degradation type such as denoising, deblurring, dehazing, deraining, or low-light enhancement. While task-specific models have achieved impressive results in their respective domains, they typically lack generalization capability.
Such single-purpose models are difficult to scale in practical applications, as switching between restoration tasks often requires retraining or task-specific adaptation, resulting in high deployment overhead and limited flexibility.
This has spurred increasing interest in All-in-One Image Restoration (AiOIR), which aims to develop unified frameworks capable of handling diverse degradations under a single model.

Recent AiOIR approaches have explored a variety of strategies to achieve generality, including task-specific architectures~\cite{li2022all}, task-specific priors~\cite{valanarasu2022transweather}, frequency-aware modulation~\cite{cui2024adair}, contrastive learning~\cite{chen2022learning}, and prompt-driven guidance~\cite{zeng2025vision}. However, these methods typically rely on predefined degradation types~\cite{potlapalli2023promptir} and rigid priors, making them less effective in handling real-world images where degradations are often compound, spatially heterogeneous, and lack explicit degradation type labels. Such limitations highlight the need for a more flexible and powerful generative framework.

\begin{figure}[t] 
    \centering
    \includegraphics[width=0.95\linewidth]{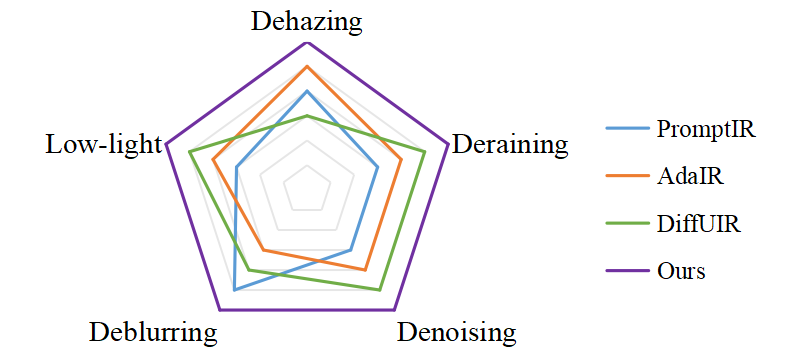} 
    \caption{
    MUSIQ-based comparison across five restoration tasks. Our method achieves consistently superior results.
    }
  \label{fig:rader_comparison}
\end{figure}

\begin{figure*}[ht]
    \centering
    \includegraphics[width=0.95 \linewidth]{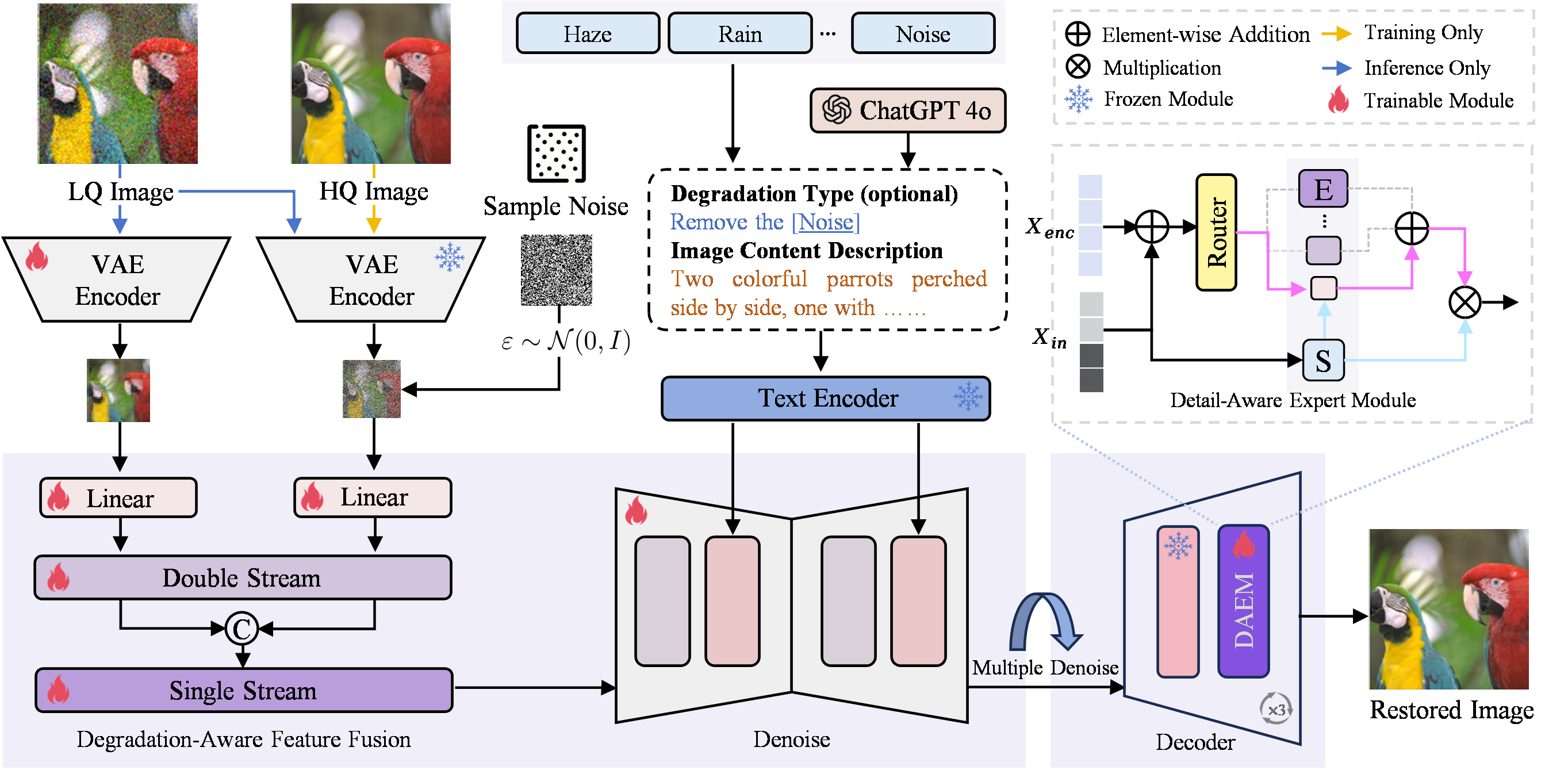}
    \caption{
     Overall architecture of the proposed UniLDiff framework.
    }
    \label{fig:pipline}
\end{figure*}

In this context, diffusion models, particularly Latent Diffusion Models (LDMs), have emerged as promising candidates for unified image restoration, owing to their strong generative priors and success in high-fidelity image synthesis and cross-modal tasks~\cite{chen2025faithdiff, jiang2024survey}. 
However, current diffusion-based AiOIR methods commonly rely on global textual or visual prompts, each with intrinsic limitations: textual prompts offer only global semantic cues and lack spatial specificity~\cite{jiang2024autodir}, while visual prompts depend on pre-trained degradation encoders or modality-specific tokens, assuming known and spatially uniform degradation types~\cite{luo2023controlling}. 
As a result, prompt-based conditioning constrains the model’s capacity to identify and respond to fine-grained degradations, limiting its adaptability despite the inherent generative strength of diffusion models.
Additionally, LDMs often suffer from detail loss due to the high compression of VAE encoders~\cite{kingma2013auto} and the progressive nature of iterative sampling~\cite{dai2023emu}, resulting in blurred textures and incomplete structures.

To address these challenges, we propose UniLDiff, a novel LDM-based unified image restoration framework that explicitly integrates degradation-aware and detail-aware mechanisms into the diffusion process. Specifically, we introduce the Degradation-Aware Feature Fusion (DAFF) module into the early layers of the diffusion UNet to inject low-quality (LQ) features into every denoising step. By leveraging decoupled fusion and adaptive modulation, DAFF provides dynamic, spatially adaptive guidance, enabling the model to perceive and adapt to diverse and heterogeneous degradations throughout the generative trajectory, without relying on predefined degradation assumptions. Furthermore, we design a Detail-Aware Expert Module (DAEM) within the decoder, which leverages decoder features and skip-connected encoder cues to dynamically activate expert branches, thereby enhancing high-frequency details and suppress structural artifacts.
A preview of our results is shown in Figure~\ref{fig:rader_comparison}. UniLDiff consistently outperforms previous methods across five representative tasks in terms of perceptual quality, demonstrating its robustness and generalization in unified multi-task restoration.
 
Our contributions can be summarized as follows:
\begin{itemize}
    \item We propose UniLDiff, a unified LDM-based framework that integrates restoration priors into the denoising process for flexible and high-fidelity image restoration.

    \item We design the DAFF that performs step-wise alignment between low-quality features and evolving latent variables, enabling the model to implicitly adapt to diverse degradations without predefined labels.

    \item We develop the DAEM that dynamically routes decoder features through expert branches guided by skip-connected encoder features, enabling targeted restoration of fine textures and structural details.

    \item Extensive experiments across multi-task and composite restoration benchmarks demonstrate that UniLDiff achieves superior perceptual quality and generalization compared to state-of-the-art methods. 
\end{itemize}

\section{Related Work}

\subsection{Non-Diffusion-Based All-in-One Image Restoration}
Traditional AiOIR methods are primarily built upon Convolutional Neural Networks (CNNs) or Transformer architectures, aiming to improve adaptability to diverse degradation types through the incorporation of task-specific structures and priors.
In recent years, various strategies have been proposed to enhance model generalization, including task-specific architectural designs~\cite{li2022all}, explicit prior modeling~\cite{valanarasu2022transweather}, frequency-aware modulation~\cite{cui2024adair}, contrastive learning~\cite{chen2022learning}, and MOE-based frameworks~\cite{zamfir2025complexity}.
Among these, prompt learning has emerged as a dominant paradigm in AiOIR. It enables task-conditioned modeling through visual prompts~\cite{zeng2025vision, tian2025degradation, cui2024adair}, textual prompts~\cite{yan2025textual}, or multimodal prompts~\cite{duan2024uniprocessor}. While these approaches improve modularity and task awareness, they often rely on predefined degradation types and rigid priors~\cite{potlapalli2023promptir}, limiting their effectiveness in real-world scenarios where degradations are compound, spatially variable, and label-agnostic.

\subsection{Diffusion-Based All-in-One Image Restoration}
Recently, diffusion models have demonstrated strong potential in AiOIR, leveraging their powerful generative priors and high-fidelity image synthesis capabilities. AutoDIR~\cite{jiang2024autodir} pioneered text-guided diffusion by using natural language prompts for restoration. MPerceiver~\cite{ai2024multimodal} extended this approach with multimodal CLIP features to handle generalized degradations. DiffUIR~\cite{zheng2024selective} further introduced a selective hourglass mapping strategy that combines strong condition guidance with shared distribution modeling for unified multi-task restoration. 
DaCLIP~\cite{luo2023controlling} leverages contrastive language-image pretraining to align degradation-aware prompts with visual features, enabling adaptive restoration under diverse degradation types.
These works demonstrate the versatility of diffusion models in AiOIR, particularly under scenarios involving mixed or unknown degradations. However, existing approaches often rely on global textual or visual prompts, which provide limited spatial specificity and typically assume known, uniform degradation types. This restricts the model’s ability to localize and respond to compound or fine-grained degradations that are common in real-world scenarios.

\section{Method}

We present UniLDiff, a unified image restoration framework that unlocks the power of diffusion priors. As illustrated in Figure~\ref{fig:pipline}, UniLDiff employs a pre-trained VAE encoder to project both low-quality (LQ) and high-quality (HQ) images into a latent space, extracting multi-scale features $f^{LQ}$ and $f^{HQ}$. The forward diffusion process adds noise to $f^{HQ}$ to produce latent variables $X_t^{HQ}$ at each timestep $t$. To inject degradation-aware guidance into the denoising trajectory, the DAFF module is integrated into the early layers of the diffusion UNet, aligning $f^{LQ}$ with $X_t^{HQ}$ at every step. Additionally, the DAEM is applied in the decoder to enhance high-frequency textures and preserve fine-grained structural details.

\subsection{Degradation-Aware Feature Fusion (DAFF)}

\begin{figure}[t] 
    \centering
    \includegraphics[width=1.0\linewidth]{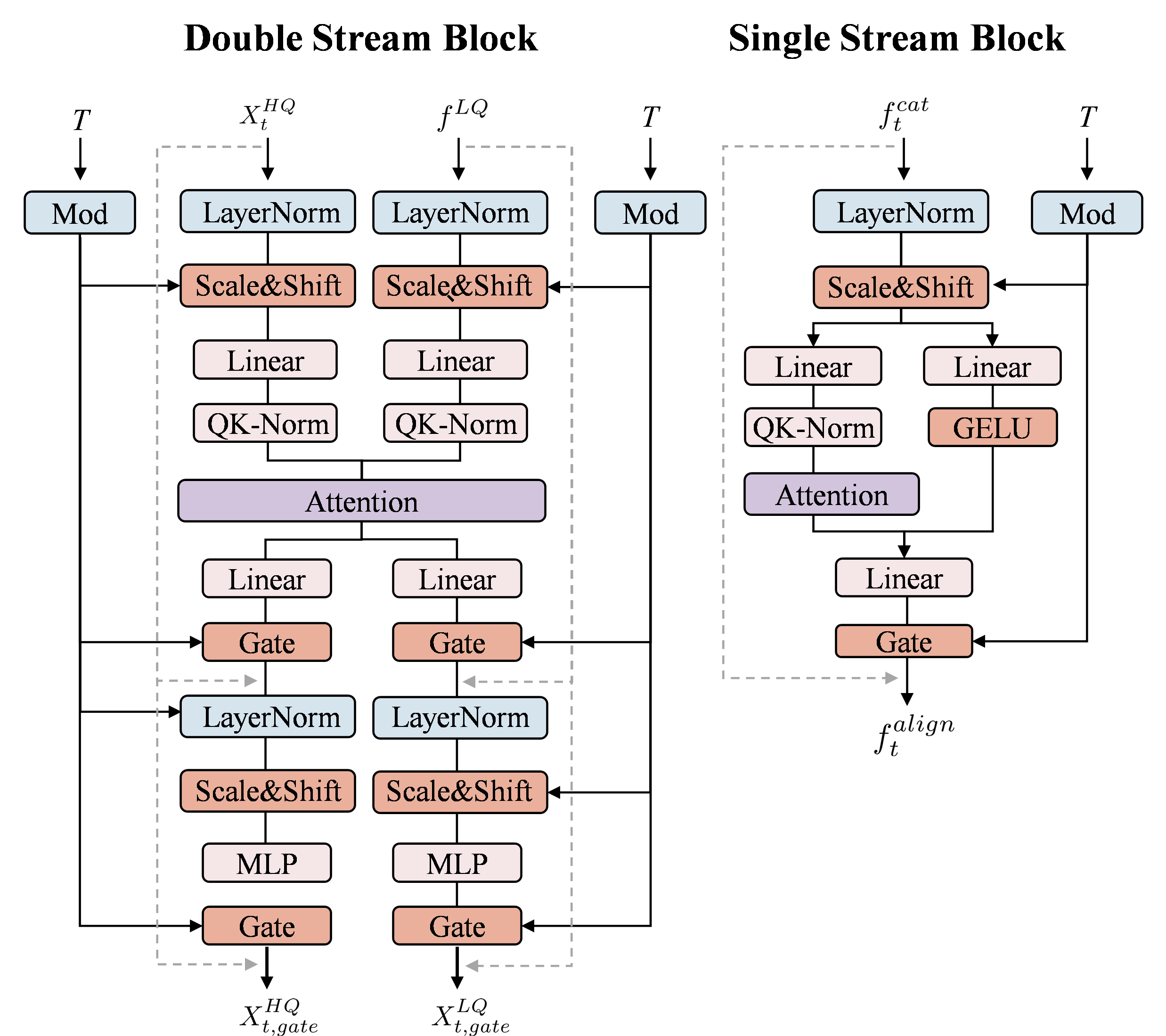} 
    \caption{Architecture of the proposed Degradation-Aware Feature Fusion (DAFF).}
  \label{fig:daff}
\end{figure}

In diffusion-based image restoration, a common practice is to simply concatenate or add the LQ features $f^{LQ}$ with the latent variable $X_t^{HQ}$ at each timestep $t$. However, this naive fusion can be suboptimal: as $X_t^{HQ}$ progressively approaches the clean HQ representation, static $f^{LQ}$ may introduce interference, undermining structural consistency in later stages.

To mitigate this, we introduce the DAFF module, which dynamically aligns LQ features with evolving latent representations at each diffusion timestep, providing degradation-aware modulation to condition the denoising trajectory according to the inferred degradation type. Inspired by FLUX~\cite{labs2025flux1kontextflowmatching}, we adopt a cascaded fusion design combining double-stream and single-stream pathways (Fig.~\ref{fig:daff}), leveraging joint and disentangled feature modeling.

\subsubsection{Double Stream Block.}
To capture degradation-specific cues and maintain structural disentanglement, we first process the $f^{LQ}$ and $X_t^{HQ}$ through separate branches. Each branch independently undergoes LayerNorm and conditional modulation before computing its own query, key, and value matrices:
\begin{align}
Q_t^D, K_t^D, V_t^D &= \text{concat}[{QKV}(f^{LQ}), {QKV}(X_{t}^{HQ})] \\
A_t^D &= \text{softmax}\left(\frac{Q_t^D (K_t^D)^T + PE}{\sqrt{d}}\right) V_t^D 
\end{align}

The attention output $A_t^D$ is projected back via a gating mechanism to obtain $X_{t,gate}^{LQ}$ and $X_{t,gate}^{HQ}$. This design enables bidirectional interaction between streams, where $X_t^{HQ}$ receives degradation-aware guidance from $f^{LQ}$ while preserving its own structural priors. Compared to naive fusion, this setup enhances alignment and reduces structural artifacts.

\subsubsection{Single Stream Block.}
The gated features are concatenated as $f_t^{cat} = \text{concat}[X_{t,gate}^{LQ}, X_{t,gate}^{HQ}]$, then normalized and linearly projected to produce the attention triplet and an auxiliary vector. The attention is computed as:
\begin{align}
[Q_t^S, K_t^S, V_t^S], M = \text{Linear}_1(f_t^{{cat}})\\
A_t^S = \text{softmax}\left(\frac{Q_t^S (K_t^S)^T + PE}{\sqrt{d}}\right) V_t^S
\end{align}
where PE is positional embeddings. We concatenate $A_t^S$ with $\phi(M)$ and feed it through a second linear layer. The output is combined with the input via gating to yield $f_t^{align}$:
\begin{align}
f_t^{align} = f^{cat}_t + g \cdot \text{Linear}_2(A_t^S, \phi(M))
\end{align}
where $\phi(\cdot)$ is a nonlinear activation and $g$ controls fusion strength.
This single-stream fusion enhances feature coherence and allows fine-grained modulation of degradation influence, mitigating guidance mismatch and improving structural consistency.

To enhance semantic alignment, we also incorporate a content embedding $c$ from a pre-trained text encoder, optionally including a task prompt, and fuse it with $f_t^{align}$ via a lightweight cross-attention mechanism. In summary, given the latent HQ feature $X_t^{HQ}$ at diffusion step $t$, the degradation feature $f^{LQ}$, and the content embedding $c$, the model predicts the latent HQ feature $X_{t-1}^{HQ}$ at the previous timestep as follows:
\begin{align}
x_{t-1}^{HQ} = \frac{1}{\sqrt{\alpha_t}} \left(x_t^{HQ} - \frac{1-\alpha_t}{\sqrt{1-\bar{\alpha}_t}} \hat{\epsilon}_\theta(f_t^{align}, c, t)\right) + \sigma_t z
\end{align}
where $\hat{\epsilon}_\theta$ is the predicted noise from the denoising UNet, conditioned on $f_t^{align}$, $c$, and $t$. The variable $z \sim \mathcal{N}(0, I)$ adds stochasticity, while $\alpha_t$, $\sigma_t$, and $\bar{\alpha}_t = \prod_{i=1}^{t} \alpha_i$ are diffusion schedule parameters controlling denoising and noise levels.

\subsection{Detail-Aware Expert Module (DAEM)}

\begin{figure}[t] 
    \centering
    \includegraphics[width=0.9\linewidth]{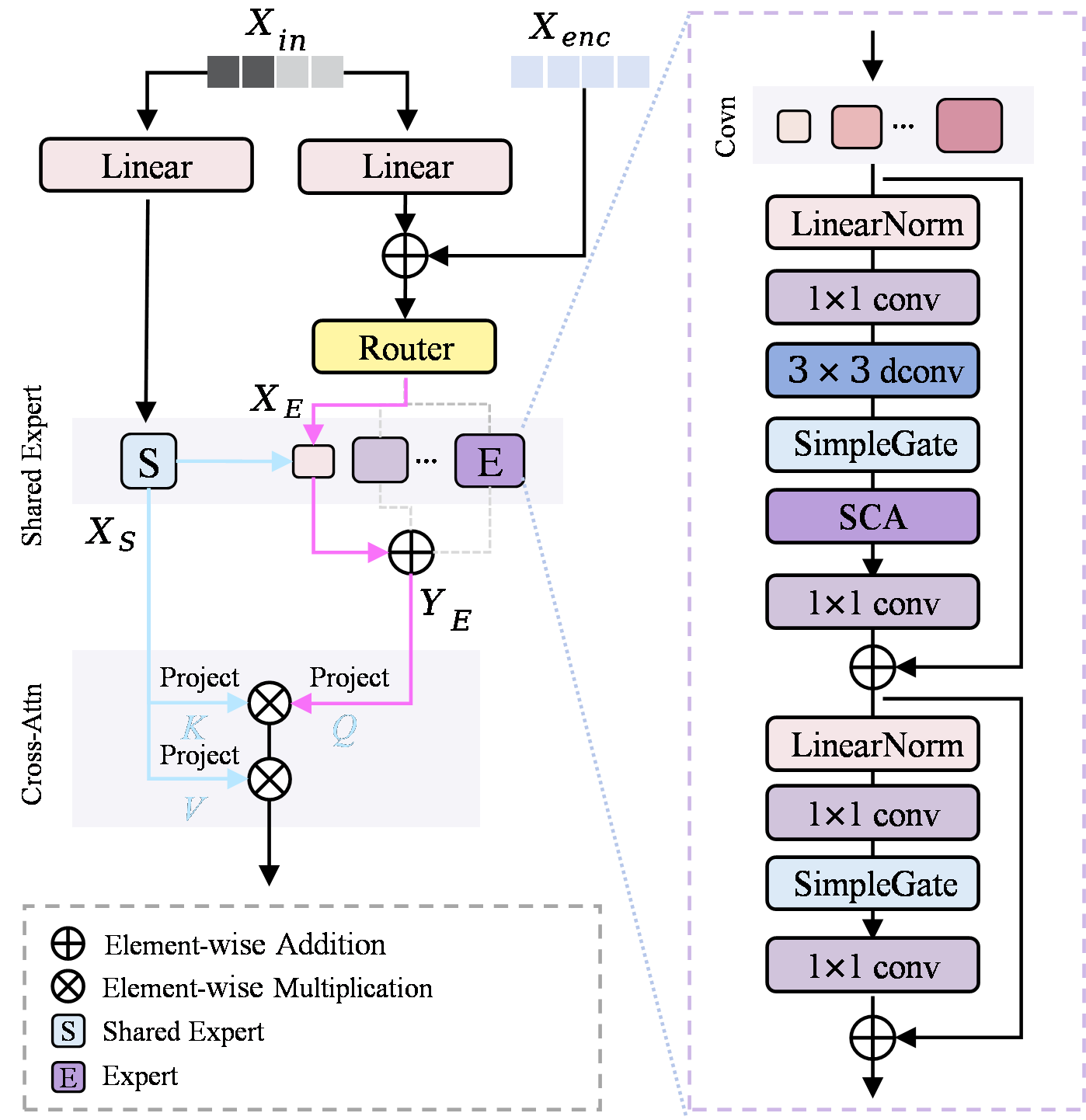} 
    \caption{
    Architecture of the DAEM, which uses experts with varied receptive fields to adaptively refine details from skip-connected encoder features.
    }

  \label{fig:daem}
\end{figure}

Although the DAFF module improves degradation perception and adaptation, LDMs still struggle to recover fine details, often resulting in texture blurring and structural loss in small objects. These limitations arise from the high compression ratio of the VAE encoder, which discards high-frequency signals, and the progressive sampling process, which introduces texture hallucinations~\cite{dai2023emu,zhu2023designing}. Furthermore, in multi-degradation scenarios, spatially heterogeneous degradations challenge the expressiveness of a single reconstruction path.

To mitigate these issues, we introduce the DAEM—a specialized decoder component designed to selectively enhance high-frequency structures and texture fidelity across spatially diverse regions (see Fig.~\ref{fig:daem}). Unlike conventional decoders that operate uniformly, DAEM leverages a Mixture-of-Experts (MoE) architecture to dynamically specialize its response based on local degradation context. More importantly, DAEM integrates encoder features via skip connections to retrieve early-stage, high-resolution cues that are otherwise lost in the VAE compression process. By fusing these fine-grained encoder features with the decoder’s latent representations, DAEM explicitly injects detail priors into the reconstruction pipeline—enabling targeted enhancement of textures, edges, and small structures.

Specifically, for each input $x$, a lightweight routing function computes the expert assignment as:
\begin{equation}
    \text{Router}(x) = \text{top-}k \left( \text{Softmax}(W x + \boldsymbol{\xi}) \right),
\end{equation}
where $W$ denotes the router weight, $\boldsymbol{\xi} \sim \mathcal{N}(0, \sigma^2)$ is sampled Gaussian noise, and $k$ is the number of activated experts. In our experiments, we empirically set $k = 1$. Each expert $E^i$ is built using lightweight NAFBlocks~\cite{chen2022simple} with varied receptive fields, allowing for multi-scale perception of both fine and coarse structures.

To ensure global semantic coherence, a shared expert branch $S(\cdot)$ with transposed self-attention~\cite{zamir2022restormer} is added. The output of each expert is modulated by this global branch as:
\begin{equation}
    \hat{y}_E^i = E^i(x) \otimes S(x),
\end{equation}
where $\otimes$ denotes element-wise multiplication. This design effectively fuses localized, detail-specific refinement with global contextual guidance.

By bridging early encoder features and dynamic expert processing, DAEM equips the model with robust fine-detail recovery capabilities and enhanced adaptability to complex, spatially variant degradations.

\begin{figure*}[t] 
    \centering
    \includegraphics[width=0.96\linewidth]{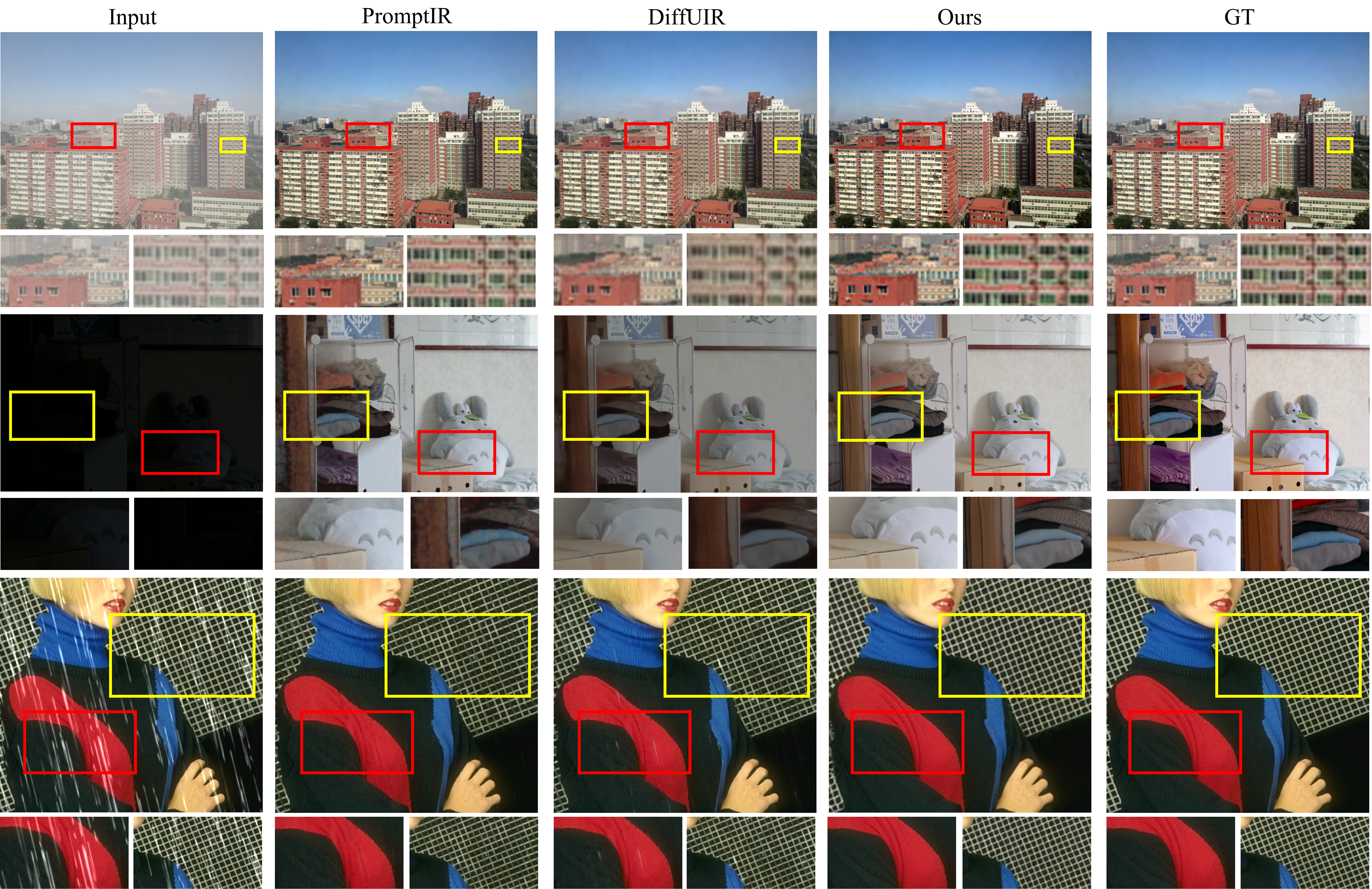} 
    \caption{
\textbf{Visual results.} In dehazing, our approach produces more natural color restoration and sharper details, in some cases surpassing the visual quality of the ground truth. For low-light enhancement, it better preserves structural edges and surface textures while avoiding over-smoothing. In deraining, our method reconstructs continuous grid structures that other methods fail to recover, demonstrating superior fidelity and generalization across diverse degradations.
}

  \label{fig:result}
\end{figure*}

\subsection{Unified Training Strategy}

Our framework comprises five core components: a pre-trained VAE encoder, a trainable VAE encoder for extracting degradation features, the DAFF module, a diffusion-based denoising network $\epsilon_\theta$, and a VAE decoder integrated with the DAEM. In this work, we adopt Stable Diffusion XL~\cite{podell2023sdxl} as our underlying latent diffusion model, leveraging its pre-trained components to initialize the encoder, decoder, and denoising UNet. The training procedure is organized into two stages: (1) degradation modeling with DAFF, and (2) detail refinement with DAEM.

\subsubsection{Stage 1: Degradation Modeling}

We first freeze the VAE encoder and denoising UNet, and train only the DAFF module. This allows DAFF to learn how to align LQ features $f^{LQ}$ with the noisy latent variable $x_t^{HQ}$ at each timestep, capturing degradation-aware priors to guide the denoising process.
Once DAFF converges, we unfreeze the trainable LQ encoder and jointly optimize it with DAFF and the denoising network $\epsilon_\theta$ to improve coordination among feature extraction, fusion, and noise prediction.
The training loss follows the standard diffusion noise estimation objective:
\begin{equation}
\mathcal{L}_{\text{stage-1}} = \left \| \epsilon - \hat { \epsilon } _ \theta \left ( \sqrt { \bar { \alpha } _ t x _ 0 ^ { H Q } } + \sqrt { 1 - \bar { \alpha } _ t } \epsilon , f ^ { L Q } , c , t \right ) \right \| _ 1 ,
\end{equation}
where $\epsilon \sim \mathcal{N}(0, I)$ is Gaussian noise, and $c$ is the auxiliary text embedding.

\subsubsection{Stage 2: Detail Refinement}

After the degradation modeling stage, we fine-tune the VAE decoder together with the DAEM to enhance the recovery of fine-grained details. This phase is supervised by a composite loss that combines pixel-wise reconstruction loss $\mathcal{L}_{\text{recon}}$ and structural similarity loss $\mathcal{L}_{\text{ssim}}$, ensuring both low-level accuracy and structural consistency.
To fully leverage the MoE architecture and prevent expert collapse, we introduce an auxiliary load-balancing loss $\mathcal{L}_{\text{aux}}$~\cite{riquelme2021scaling}, which encourages uniform expert utilization across batches. Details of its formulation are provided in the Appendix.

The total objective for this stage is:
\begin{equation}
\mathcal{L}_{\text{stage-2}} = \mathcal{L}_{\text{recon}} + \lambda_1 \mathcal{L}_{\text{ssim}} + \lambda_2 \mathcal{L}_{\text{aux}},
\end{equation}
where $\lambda_1$ and $\lambda_2$ are hyperparameters controlling the trade-off between structural preservation and expert diversity.

\begin{table*}[t]
\centering
\small
\setlength{\tabcolsep}{3pt} 
\begin{tabular}{clccccccccc}
\toprule
 & \textbf{Method} & \textbf{Venue} & \textbf{PSNR$\uparrow$} & \textbf{SSIM$\uparrow$} & \textbf{LPIPS$\downarrow$} & \textbf{DISTS$\downarrow$} & \textbf{CLIPIQA$\uparrow$} & \textbf{NIQE$\downarrow$} & \textbf{MUSIQ$\uparrow$} & \textbf{MANIQA$\uparrow$} \\
\midrule
\multirow{7}{*}{\rotatebox[origin=c]{90}{\textbf{Non-Diff}}}
& AirNet    & CVPR2022  & 31.11 & 0.9068 & 0.0925 & 0.0950 & 0.6405 & \textcolor{second}{\textbf{3.3911}} & 66.98 & 0.6470 \\
& IDR       & CVPR2023  & 30.72 & 0.8960 & 0.0949 & 0.0920 & 0.6613 & 3.5283 & 66.63 & 0.6443 \\
& PromptIR  & NeurIPS23 & 32.18 & 0.9124 & 0.0859 & 0.0864 & 0.6409 & 4.0061 & 67.21 & 0.6569 \\
& VLU-Net   & CVPR2025  & 32.55 & \textcolor{second}{\textbf{0.9157}} & 0.0796 & 0.0785 & 0.6433 & 3.6559 & 67.09 & \textcolor{second}{\textbf{0.6698}} \\
& DFPIR     & CVPR2025  & \textcolor{second}{\textbf{32.75}} & \textcolor{first}{\textbf{0.9162}} & 0.0758 & 0.0758 & \textcolor{second}{\textbf{0.6626}} & 3.5938 & 67.34 & 0.6679 \\
& AdaIR     & ICLR2025  & \textcolor{first}{\textbf{32.98}} & 0.9155 & 0.0825 & 0.0820 & 0.6435 & 3.6464 & 67.50 & 0.6685 \\
\midrule
\multirow{3}{*}{\rotatebox[origin=c]{90}{\textbf{Diff}}}
& DA-CLIP   & ICLR2024  & 30.27 & 0.8780 & \textcolor{second}{\textbf{0.0664}} & \textcolor{second}{\textbf{0.0650}} & 0.6580 & \textcolor{first}{\textbf{3.2783}} & 67.31 & 0.6663 \\
& DiffUIR & CVPR2024  & 31.89 & 0.9010 & 0.0959 & 0.0964 & 0.6163 & 3.7371 & \textcolor{second}{\textbf{67.51}} & 0.6570 \\
& \textbf{Ours }     
& -         
& 32.18 
& 0.9105 
& \textcolor{first}{\textbf{0.0651}} 
& \textcolor{first}{\textbf{0.0639}} 
& \textcolor{first}{\textbf{0.6653}} 
& 3.6022 
& \textcolor{first}{\textbf{68.89}} 
& \textcolor{first}{\textbf{0.7038}}
 \\
\bottomrule
\end{tabular}
\caption{Quantitative comparison on average performance across three restoration tasks: dehazing, deraining, and Gaussian denoising ($\sigma = 15$, $25$, $50$). 
Red and blue indicate the best and second-best results, respectively.
}
\label{tab:three-task}
\end{table*}

\section{Experiment}

\begin{table*}[ht]
\centering
\small
\setlength{\tabcolsep}{2pt} 
\begin{tabular}{clcccccccccc}
\toprule
\multirow{2}{*}{} & \multirow{2}{*}{\textbf{Method}} 
& \multicolumn{2}{c}{\textbf{Dehazing}} 
& \multicolumn{2}{c}{\textbf{Deraining}} 
& \multicolumn{2}{c}{\textbf{Denoising}} 
& \multicolumn{2}{c}{\textbf{Deblurring}} 
& \multicolumn{2}{c}{\textbf{Low-light}} \\
& & MUSIQ & MANIQA & MUSIQ & MANIQA & MUSIQ & MANIQA & MUSIQ & MANIQA & MUSIQ & MANIQA \\
\midrule
\multirow{5}{*}{\rotatebox[origin=c]{90}{\small\textbf{Non-Diff}}}
& PromptIR     & 65.43 & 0.6829 & 68.00 & 0.6679 & 67.29 & 0.6295 & 36.60 & 0.4250 & 64.76 & 0.5922 \\
& InstructIR   & 59.75 & 0.6622 & 68.01 & 0.6659 & 68.35 & 0.6598 & \textcolor{second}{\textbf{38.83}} & 0.4007 & 67.34 & 0.6061 \\
& AdaIR        & \textcolor{second}{\textbf{65.83}} & \textcolor{second}{\textbf{0.6923}} & 68.31 & 0.6720 & 67.65 & 0.6593 & 33.41 & 0.3763 & 69.43 & 0.6309 \\
& VLU-Net      & 65.81 & 0.6920 & 68.45 & \textcolor{second}{\textbf{0.6740}} & \textcolor{second}{\textbf{68.38}} & \textcolor{second}{\textbf{0.6669}} & 32.48 & 0.3568 & 65.21 & 0.5948 \\
& DFPIR        & 65.61 & 0.6898 & 68.29 & 0.6700 & 68.06 & 0.6563 & 36.93 & 0.3978 & 69.19 & 0.6234 \\
\midrule
\multirow{3}{*}{\rotatebox[origin=c]{90}{\small\textbf{Diff}}}
& DA-CLIP      & \textcolor{first}{\textbf{67.88}} & 0.6524 & \textcolor{second}{\textbf{68.60}} & 0.6695 & 67.88 & 0.6524 & 36.89 & \textcolor{second}{\textbf{0.4298}} & \textcolor{second}{\textbf{72.59}} & \textcolor{first}{\textbf{0.6619}} \\
& DiffUIR      & 64.84 & 0.6864 & 68.40 & 0.6682 & 68.19 & 0.6383 & 33.63 & 0.3529 & 71.13 & 0.6246 \\

& \textbf{Ours} 
& 66.42 
& \textcolor{first}{\textbf{0.6948}} 
& \textcolor{first}{\textbf{68.69}} 
& \textcolor{first}{\textbf{0.6903}} 
& \textcolor{first}{\textbf{68.49}} 
& \textcolor{first}{\textbf{0.6700}} 
& \textcolor{first}{\textbf{38.91}} 
& \textcolor{first}{\textbf{0.4440}} 
& \textcolor{first}{\textbf{72.77}} 
& \textcolor{second}{\textbf{0.6528}} \\
\bottomrule
\end{tabular}
\caption{
Quantitative comparison of no-reference metrics MUSIQ~($\uparrow$) and MANIQA~($\uparrow$) across five restoration tasks: dehazing, deraining, denoising ($\sigma = 25$), deblurring, and low-light enhancement. Red and blue indicate the best and second-best results.
}
\label{tab:five-task}
\end{table*}

\begin{table*}[ht]
\centering
\small
\setlength{\tabcolsep}{3pt}
\begin{tabular}{l cc cc cc cc cc cc cc}
\toprule
\multirow{2}{*}{\textbf{Method}} 
& \multicolumn{2}{c}{\textbf{L+H}} 
& \multicolumn{2}{c}{\textbf{L+R}} 
& \multicolumn{2}{c}{\textbf{L+S}} 
& \multicolumn{2}{c}{\textbf{H+R}} 
& \multicolumn{2}{c}{\textbf{H+S}} 
& \multicolumn{2}{c}{\textbf{L+H+R}} 
& \multicolumn{2}{c}{\textbf{L+H+S}} \\
& PSNR & SSIM & PSNR & SSIM & PSNR & SSIM 
& PSNR & SSIM & PSNR & SSIM 
& PSNR & SSIM & PSNR & SSIM \\
\midrule
AirNet    
& 23.23 & 0.779 & 23.21 & 0.768 & 23.06 & 0.763 
& 25.04 & 0.883 & 25.07 & 0.884 
& 22.85 & 0.751 & 22.61 & 0.748 \\
PromptIR  
& 24.49 & 0.789 & 24.33 & 0.773 & 24.12 & 0.770 
& 27.93 & 0.939 & 28.02 & 0.940 
& 23.87 & 0.763 & 23.55 & 0.759 \\
WGWSNet   
& 23.95 & 0.772 & 23.70 & 0.764 & 23.54 & 0.759 
& 27.53 & 0.890 & 27.60 & 0.891 
& 23.30 & 0.748 & 23.07 & 0.745 \\
WeatherDiff 
& 22.36 & 0.756 & 22.25 & 0.750 & 22.10 & 0.745 
& 23.82 & 0.867 & 23.91 & 0.869 
& 21.87 & 0.732 & 21.60 & 0.730 \\
OneRestore  
& 25.79 & \textcolor{first}{\textbf{0.822}} & 25.65 & \textcolor{first}{\textbf{0.811}} & 25.44 & \textcolor{first}{\textbf{0.808}} 
& 29.81 & 0.960 & 29.92 & 0.961 
& 25.23 & \textcolor{first}{\textbf{0.797}} & 24.93 & \textcolor{first}{\textbf{0.793}} \\
MoCE-IR-S       
& \textcolor{second}{\textbf{26.24}} & 0.817 & \textcolor{second}{\textbf{26.05}} & 0.804 & \textcolor{second}{\textbf{26.04}} & 0.802 
& \textcolor{second}{\textbf{29.93}} & \textcolor{second}{\textbf{0.961}} & \textcolor{second}{\textbf{30.19}} & \textcolor{second}{\textbf{0.962}} 
& \textcolor{second}{\textbf{25.41}} & 0.789 & \textcolor{second}{\textbf{25.39}} & 0.787 \\
\textbf{Ours}                    
& \textcolor{first}{\textbf{26.50}} & \textcolor{second}{\textbf{0.820}} & \textcolor{first}{\textbf{26.40}} & \textcolor{second}{\textbf{0.808}} & \textcolor{first}{\textbf{26.10}} & \textcolor{second}{\textbf{0.805}} 
& \textcolor{first}{\textbf{30.10}} & \textcolor{first}{\textbf{0.963}} & \textcolor{first}{\textbf{30.50}} & \textcolor{first}{\textbf{0.964}} 
& \textcolor{first}{\textbf{25.80}} & \textcolor{second}{\textbf{0.794}} & \textcolor{first}{\textbf{25.50}} & \textcolor{second}{\textbf{0.790}} \\
\bottomrule
\end{tabular}
\caption{
Comparison on composite degradation types. PSNR~($\uparrow$) and SSIM~($\uparrow$) are reported. L: Low, H: Haze, R: Rain, S: Snow. Red and blue indicate the best and second-best results, respectively.
}
\label{tab:cdd11}
\end{table*}

Following prior work in general-purpose image restoration ~\cite{jiang2024survey}, we evaluate our model under two distinct settings: (i) Universal Degradation Restoration, where a single model is trained to handle multiple degradation types, including combinations of three and five types; and (ii) Compositional Degradation Restoration, where the model is assessed on both individual and mixed degradation scenarios (up to three types) using the same unified architecture.

\subsection{Experimental Settings}

\subsubsection{Datasets.}
We aggregate a diverse set of widely-used datasets across different degradation types. For denoising, we use BSD400~\cite{arbelaez2010contour} and WED~\cite{ma2016waterloo} with synthetic Gaussian noise ($\sigma=15$, $25$, $50$) for training, and BSD68~\cite{martin2001database} for testing. Rain100L~\cite{yang2017deep}, RESIDE-SOTS~\cite{li2018benchmarking}, GoPro~\cite{nah2017deep}, and LOL-v1~\cite{wei2018deep} are used for deraining, dehazing, deblurring, and low-light enhancement, respectively. To support unified modeling, we construct multi-task settings with three and five degradation types, denoted as AIO-3 and AIO-5. For compositional degradations, we additionally evaluate on CDD11~\cite{guo2024onerestore}.

\subsubsection{Metrics.}
To comprehensively evaluate restoration performance, we adopt a suite of image quality assessment metrics covering both distortion-based fidelity and perceptual quality. For full-reference evaluation, PSNR and SSIM~\cite{wang2004image} are used to assess pixel-level and structural similarity. To better align with human perception, we employ LPIPS~\cite{zhang2018unreasonable} and DISTS~\cite{ding2020image}. Additionally, to support reference-free evaluation in real-world scenarios, we include no-reference metrics such as NIQE~\cite{zhang2015feature}, MANIQA~\cite{yang2022maniqa}, MUSIQ~\cite{ke2021musiq}, and CLIPIQA~\cite{wang2023exploring}.

\subsection{Comparison with State-of-the-Arts}

\subsubsection{Universal Degradation Restoration.}

To validate the generalization and restoration performance of our method across diverse degradation types, we first conduct systematic comparisons under a three-task joint training setup involving dehazing, deraining, and denoising. As shown in Table~\ref{tab:three-task}, although our PSNR and SSIM are slightly lower than some state-of-the-art non-diffusion methods (e.g., DFPIR, AdaIR), our method achieves the best scores on perceptual metrics, including LPIPS, DISTS, and CLIPIQA. Notably, we obtain the highest MUSIQ and MANIQA scores, reflecting superior perceptual quality. Compared with other diffusion-based methods, our model consistently outperforms them across all metrics, demonstrating enhanced fidelity and perceptual performance.

We further extend the evaluation to a five-task setup by including deblurring and low-light enhancement, where task conflict becomes more pronounced. As shown in Table~\ref{tab:five-task}, our method achieves the top MUSIQ score in most tasks and ranks among the best in MANIQA, confirming its robustness and adaptability under complex degradation scenarios. In contrast, other diffusion-based methods suffer from inconsistent task-wise performance. Figure~\ref{fig:result} presents representative visual comparisons across five distinct restoration tasks, further demonstrating the capability of our method to recover fine structural details under a unified multi-task setting. More detailed results including runtime and parameter statistics are provided in Appendix.

\subsubsection{Composited Degradations.}

To better simulate real-world restoration demands, we adopt a unified setting based on the CDD11 dataset, which includes 11 degradation types spanning haze, rain, snow, low-light, and their various combinations. This comprehensive setup imposes higher demands on generalization and robustness. As shown in Table~\ref{tab:cdd11}, our method consistently delivers strong performance across all tasks. We often rank first or second in PSNR and SSIM, achieving state-of-the-art results on two-fold composite degradations such as L+H, L+S, and H+S. Even under challenging three-fold degradations (L+H+R and L+H+S), our model maintains leading performance with PSNR/SSIM of 25.80/0.794 and 25.50/0.790, respectively—demonstrating its robust cross-degradation modeling capability.
On average, our approach achieves 29.35 dB PSNR and 0.886 SSIM across all 11 tasks, outperforming the representative MoCE-IR baseline by 0.30 dB and 0.005. While Table~\ref{tab:cdd11} reports only objective metrics, our method also excels in perceptual quality and visual consistency, further validating its unified modeling capabilities under complex degradation.

\subsection{Ablation Study}

\subsubsection{Structural Design of the DAFF}
To evaluate the necessity of the proposed design in the DAFF module, we conduct a comprehensive ablation study across four configurations: (1) no DAFF fusion (baseline); (2) single-stream fusion only; (3) double-stream fusion only; and (4) our full DAFF design combining both paths. In addition, we compare our method with a representative residual self-attention (RSA) fusion strategy~\cite{radford2021learning}, which replaces DAFF with a standard attention block.

As reported in Table~\ref{tab:ablation_daff}, both single-stream and double-stream branches offer partial benefits—single-stream fusion improves integration but lacks representation disentanglement, while double-stream enhances alignment but suffers from fusion instability. Our full DAFF achieves the best performance by combining their respective strengths.
Moreover, RSA performs worse than our method due to its lack of timestep-aware modulation. While RSA captures spatial attention, it fails to adaptively adjust the influence of degradation cues throughout the diffusion process. In contrast, DAFF leverages timestep $t$ as a dynamic conditioning signal, enabling consistent and adaptive fusion across evolving latent states.

\begin{table}[t]
\centering
\small
\setlength{\tabcolsep}{4pt}
\begin{tabular}{lcc}
\toprule
\textbf{Fusion Strategy} & \textbf{PSNR}~$\uparrow$ & \textbf{MUSIQ}~$\uparrow$ \\
\midrule
No fusion & 23.12 & 41.31 \\
Single-stream only & 25.39 & 49.26 \\
Double-stream only & 25.84 & 51.02 \\
Residual Self-Attention (RSA) & 24.08 & 48.11 \\
\textbf{Ours (Full DAFF)} & \textbf{27.14} & \textbf{61.35} \\
\bottomrule
\end{tabular}
\caption{DAFF ablation results averaged over five tasks. Fusion of double-stream and single-stream modules with timestep-aware modulation achieves the best performance.}
\label{tab:ablation_daff}
\end{table}

\subsubsection{Effectiveness of the DAFF}

To further validate the overall effectiveness of the DAFF module, we compare different configurations with and without DAFF under both task prompt-enabled and prompt-free settings. As shown in Table~\ref{tab:ablation}, incorporating task prompts alone provides semantic guidance and improves performance in globally homogeneous degradations. However, it remains insufficient for handling spatially localized degradations such as rain or shadow.
By contrast, adding DAFF enables dynamic interaction between degradation cues and evolving latent features, resulting in significantly improved restoration fidelity. Notably, the combination of DAFF and task prompt yields the best performance, demonstrating their complementarity: task prompts offer global semantics, while DAFF contributes fine-grained degradation awareness.
These results highlight that DAFF is not only effective independently, but also synergizes well with prompt-based guidance to improve robustness and quality across diverse degradation scenarios.

\begin{table}[t]
\centering
\small
\setlength{\tabcolsep}{4pt}  
\begin{tabular}{cccccc}
\toprule
\textbf{DAFF} & \textbf{Task Prompt} & \textbf{DAEM} & \textbf{PSNR}$\uparrow$ & \textbf{MUSIQ}$\uparrow$ \\
\midrule
\ding{55} & \ding{55} & \ding{55} & 23.12 & 41.31 \\
\ding{55} & \ding{51} & \ding{55} & 25.87 & 50.12 \\
\ding{51} & \ding{55} & \ding{55} & 27.14 & 61.35 \\
\ding{51} & \ding{51} & \ding{55} & 27.36 & 62.77 \\
\ding{51} & \ding{51} & \ding{51} & \textbf{30.27} & \textbf{63.06} \\
\bottomrule
\end{tabular}
\caption{Component-wise ablation study, averaged over five restoration tasks.}
\label{tab:ablation}
\end{table}

\begin{figure}[t] 
    \centering
    \includegraphics[width=0.95\linewidth]{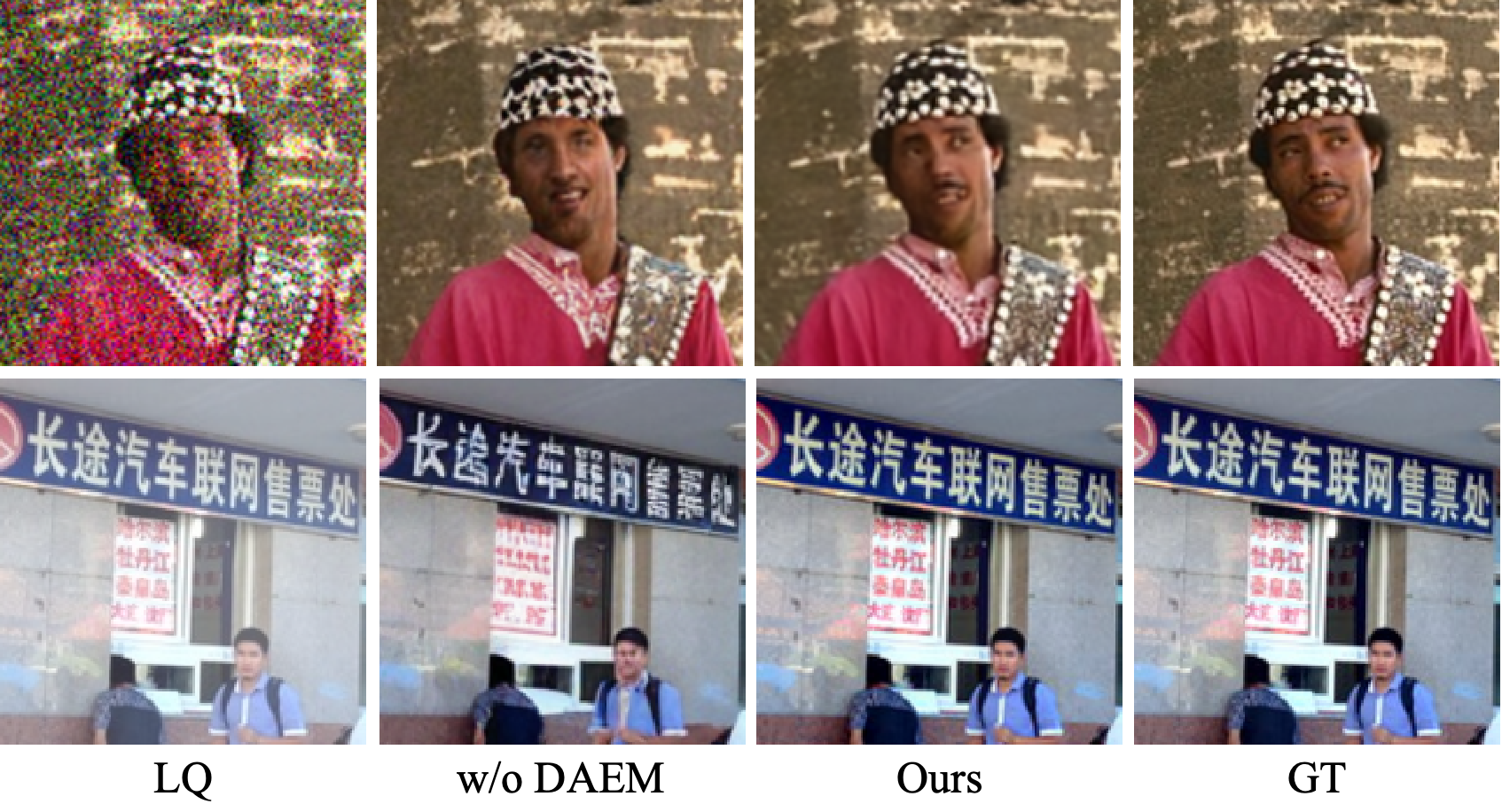} 
    \caption{
    Effectiveness of the DAEM.
    }
  \label{fig:ablation}
\end{figure}

\subsubsection{Effectiveness of the DAEM}

While DAFF improves alignment and global consistency, fine details—such as text and facial contours—are often inaccurately restored. To address this, we incorporate the DAEM, which selectively enhances high-frequency structures using a dynamic mixture-of-experts design.
Each expert branch captures distinct local patterns via diverse receptive fields, and a sparse gating mechanism routes features to the most suitable expert based on local context. As shown in Table~\ref{tab:ablation}, introducing DAEM on top of DAFF and prompt guidance yields a significant boost in PSNR and MUSIQ. Figure~\ref{fig:ablation} further shows that DAEM visibly reduces residual artifacts and sharpens fine structures. These results confirm the critical role of DAEM in refining details and complementing DAFF's structural fusion.

\section{Conclusion}

In this paper, we present a novel All-in-One image restoration framework built upon LDMs. To address the limitations of prompt-conditioned diffusion in modeling diverse degradation types, we introduce a structure-aware guidance mechanism through the DAFF module, enabling implicit adaptation to various degradations. Additionally, the proposed DAEM enhances fine-detail recovery and structural consistency in the decoding stage. Extensive experiments across multi-task and mixed degradation setting demonstrate the superiority of our method.

\bibliography{aaai2026}

\clearpage

\twocolumn[
\begin{center}
\textbf{\LARGE Appendix}
\end{center}

]

\section{Experiment}

\subsubsection*{Additional Quantitative Results on PSNR and SSIM}

\begin{figure*}[h]
    \centering
    \includegraphics[width=1 \linewidth]{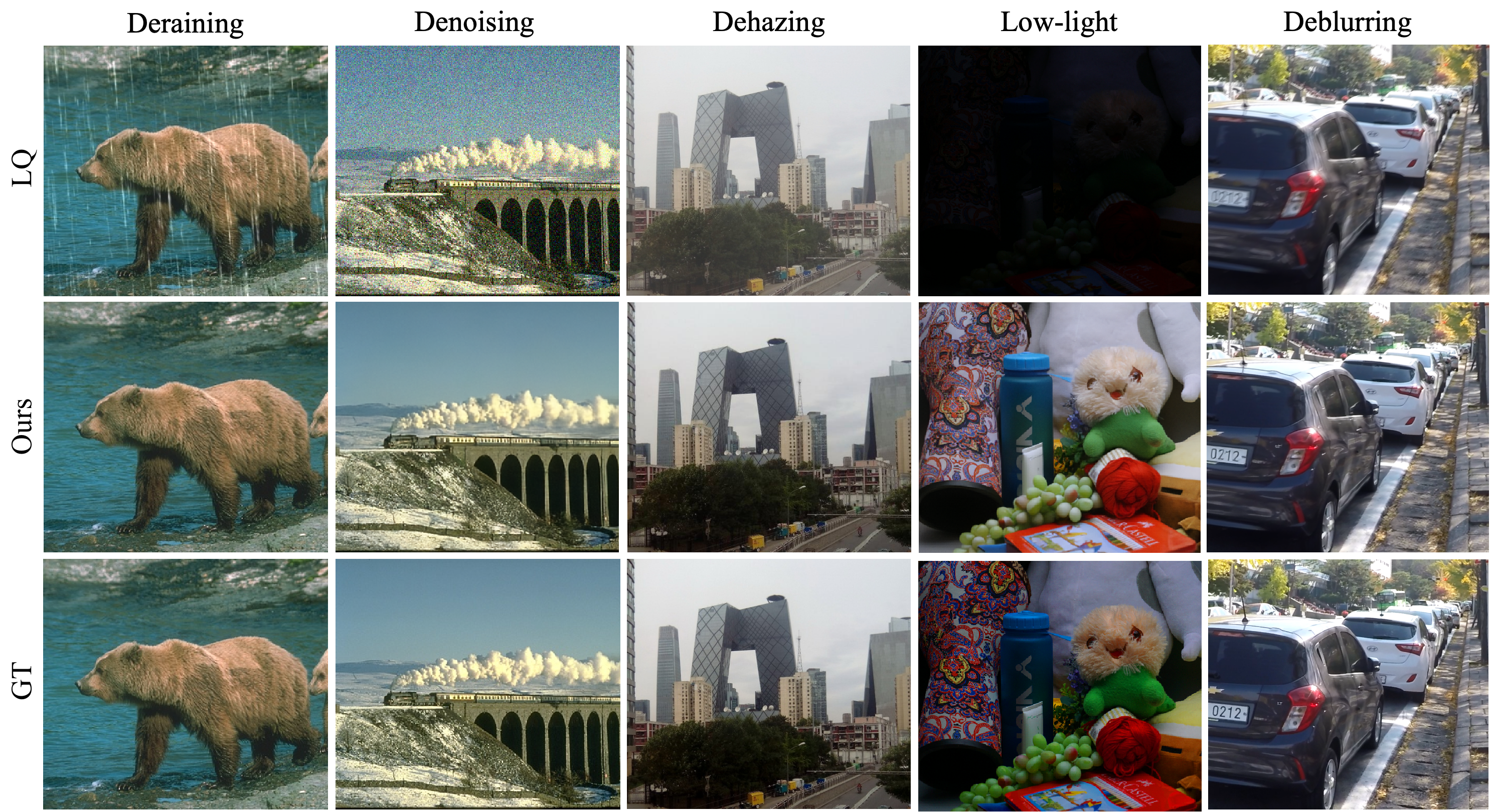}
    \caption{
     Qualitative comparison on five single degradation types: deraining, denoising, dehazing, low-light enhancement, and deblurring.
    }
    \label{fig:appendix-cmp}
\end{figure*}

\begin{table*}[t!]
\centering
\small
\setlength{\tabcolsep}{8.5pt} 
\begin{tabular}{clcccccccccc} 
\toprule
\multirow{2}{*}{} & \multirow{2}{*}{\textbf{Method}}  & \multicolumn{2}{c}{\textbf{Dehazing}} & \multicolumn{2}{c}{\textbf{Deraining}} & \multicolumn{2}{c}{\textbf{Denoising}} & \multicolumn{2}{c}{\textbf{Deblurring}} & \multicolumn{2}{c}{\textbf{Low-light}} \\
 &  & PSNR & SSIM & PSNR & SSIM & PSNR & SSIM & PSNR & SSIM & PSNR & SSIM \\
\midrule
\multirow{5}{*}{\rotatebox[origin=c]{90}{\small\textbf{Non-Diff}}}
& PromptIR     & 29.52 & 0.9730 & 36.66 & 0.9757 & 30.90 & 0.8742 & 28.71 & \textcolor{second}{\textbf{0.8812}} & 22.46 & 0.8339 \\
& InstructIR   & 24.95 & 0.8232 & 35.78 & 0.9706 & 31.35 & 0.8868 & \textcolor{first}{\textbf{29.55}} & \textcolor{first}{\textbf{0.8897}} & 22.79 & 0.8355 \\
& AdaIR        & 30.25 & 0.9765 & \textcolor{second}{\textbf{37.86}} & \textcolor{second}{\textbf{0.9804}} & 31.31 & 0.8845 & 27.03 & 0.8523 & 22.94 & \textcolor{second}{\textbf{0.8458}} \\
& VLU-Net      & 30.58 & \textcolor{second}{\textbf{0.9790}} & \textcolor{first}{\textbf{38.34}} & \textcolor{first}{\textbf{0.9818}} & \textcolor{second}{\textbf{31.40}} & \textcolor{second}{\textbf{0.8870}} & 27.48 & 0.8464 & 22.33 & 0.8343 \\
& DFPIR        & \textcolor{first}{\textbf{31.51}} & \textcolor{first}{\textbf{0.9791}} & 37.61 & 0.9788 & 31.26 & 0.8845 & \textcolor{second}{\textbf{28.80}} & 0.8770 & \textcolor{second}{\textbf{23.66}} & 0.8444 \\
\midrule
\multirow{3}{*}{\rotatebox[origin=c]{90}{\small\textbf{Diff}}}
& DA-CLIP      & 28.98 & 0.8298 & 35.94 & 0.9701 & 28.98 & 0.8298 & 25.83 & 0.8318 & 21.10 & 0.8393 \\
& DiffUIR      & 28.97 & 0.9297 & 36.47 & 0.9757 & 31.05 & 0.8768 & 26.40 & 0.8280 & 20.19 & 0.8319 \\

& \textbf{Ours} 
& \textcolor{second}{\textbf{30.83}} 
& 0.9583 
& 37.80 
& 0.9797 
& \textcolor{first}{\textbf{31.45}} 
& \textcolor{first}{\textbf{0.8895}} 
& 27.54 
& 0.8466 
& \textcolor{first}{\textbf{23.74}} 
& \textcolor{first}{\textbf{0.8789}} \\
\bottomrule
\end{tabular}
\caption{Comparison of PSNR / SSIM across five tasks. Bold indicates the best.}
\label{tab:five-task-2}
\end{table*}

\begin{figure*}[h]
    \centering
    \includegraphics[width=1 \linewidth]{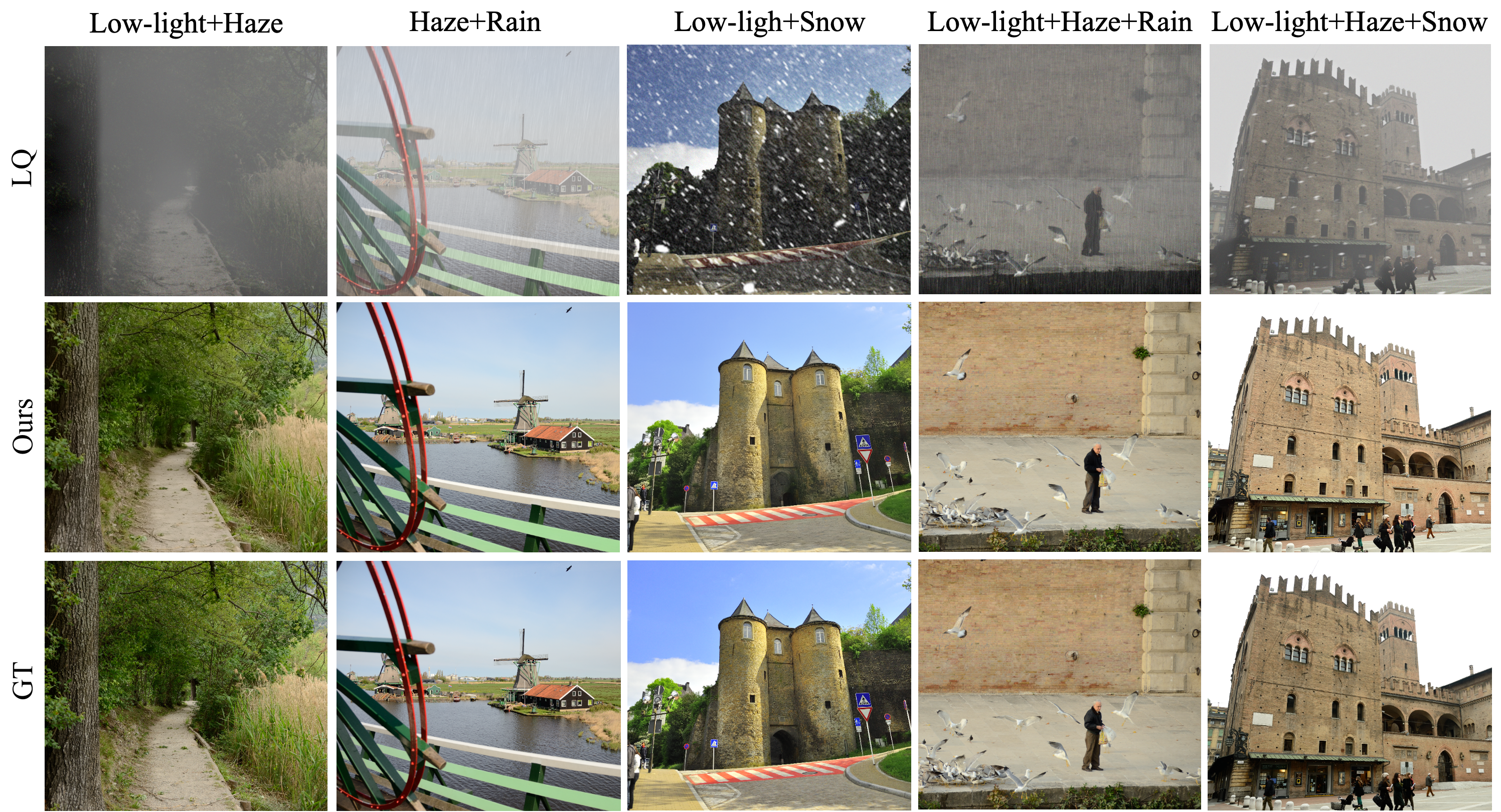}
    \caption{
     Qualitative comparison on the CDD11 dataset. Our method shows strong generalization and high-quality restoration under complex combined degradations.
    }
    \label{fig:appendix-cmp-v2}
\end{figure*}

\begin{table*}[h]
\centering
\renewcommand{\arraystretch}{1.2}
\setlength{\tabcolsep}{2pt}
\begin{adjustbox}{max width=\textwidth}
\begin{tabular}{l
cccccccccc
cccccccccccccccccccccccc}
\toprule
\multirow{2}{*}{\textbf{Method}} 
& \multicolumn{2}{c}{\textbf{L}} 
& \multicolumn{2}{c}{\textbf{H}} 
& \multicolumn{2}{c}{\textbf{R}} 
& \multicolumn{2}{c}{\textbf{S}} 
& \multicolumn{2}{c}{\textbf{L+H}} 
& \multicolumn{2}{c}{\textbf{L+R}} 
& \multicolumn{2}{c}{\textbf{L+S}} 
& \multicolumn{2}{c}{\textbf{H+R}} 
& \multicolumn{2}{c}{\textbf{H+S}} 
& \multicolumn{2}{c}{\textbf{L+H+R}} 
& \multicolumn{2}{c}{\textbf{L+H+S}} 
& \multicolumn{2}{c}{\textbf{Avg.}} \\
& PSNR & SSIM & PSNR & SSIM & PSNR & SSIM & PSNR & SSIM 
& PSNR & SSIM & PSNR & SSIM & PSNR & SSIM 
& PSNR & SSIM & PSNR & SSIM & PSNR & SSIM 
& PSNR & SSIM \\
\midrule
AirNet    
& 24.83 & .778 & 24.21 & .951 & 26.55 & .891 & 26.79 & .919 
& 23.23 & .779 & 23.21 & .768 & 23.06 & .763 & 25.04 & .883 & 25.07 & .884 
& 22.85 & .751 & 22.61 & .748 & 23.75 & .814 \\

PromptIR 
& 26.32 & .805 & 26.10 & .969 & 31.56 & .946 & 31.53 & .960 
& 24.49 & .789 & 24.33 & .773 & 24.12 & .770 & 27.93 & .939 & 28.02 & .940 
& 23.87 & .763 & 23.55 & .759 & 25.90 & .850 \\

WGWSNet 
& 24.39 & .774 & 27.90 & \textcolor{second}{\textbf{.983}} & 31.35 & .906 & 31.25 & .906 
& 23.95 & .772 & 23.70 & .764 & 23.54 & .759 & 27.53 & .890 & 27.60 & .891 
& 23.30 & .748 & 23.07 & .745 & 26.96 & .863 \\

WeatherDiff
& 23.58 & .763 & 21.99 & .943 & 24.85 & .885 & 24.80 & .883 
& 22.36 & .756 & 22.25 & .750 & 22.10 & .745 & 23.82 & .867 & 23.91 & .869 
& 21.87 & .732 & 21.60 & .730 & 22.49 & .799 \\

OneRestore 
& 26.48 & \textcolor{second}{\textbf{.826}} & 32.52 & \textcolor{first}{\textbf{.990}} & \textcolor{first}{\textbf{34.96}} & .964 & 34.31 & .973 
& 25.79 & \textcolor{first}{\textbf{.822}} & 25.65 & \textcolor{first}{\textbf{.811}} & 25.44 & \textcolor{first}{\textbf{.808}} & 29.81 & .960 & 29.92 & .961 
& 25.23 & \textcolor{first}{\textbf{.797}} & 24.93 & \textcolor{first}{\textbf{.793}} & 28.47 & .878 \\

MoCE-IR-S       
& \textcolor{second}{\textbf{27.26}} & .824 & \textcolor{second}{\textbf{32.66}} & \textcolor{first}{\textbf{.990}} & 34.31 & \textcolor{second}{\textbf{.970}} & \textcolor{second}{\textbf{35.91}} & \textcolor{second}{\textbf{.980}} 
& \textcolor{second}{\textbf{26.24}} & .817 & \textcolor{second}{\textbf{26.05}} & .804 & \textcolor{second}{\textbf{26.04}} & .802 & \textcolor{second}{\textbf{29.93}} & \textcolor{second}{\textbf{.961}} & \textcolor{second}{\textbf{30.19}} & \textcolor{second}{\textbf{.962}} 
& \textcolor{second}{\textbf{25.41}} & .789 & \textcolor{second}{\textbf{25.39}} & .787 & \textcolor{second}{\textbf{29.05}} & \textcolor{second}{\textbf{.881}} \\

Ours                    
& \textcolor{first}{\textbf{27.45}} & \textcolor{first}{\textbf{.830}} & \textcolor{first}{\textbf{32.87}} & .981 & \textcolor{second}{\textbf{34.60}} & \textcolor{first}{\textbf{.972}} & \textcolor{first}{\textbf{36.10}} & \textcolor{first}{\textbf{.983}} 
& \textcolor{first}{\textbf{26.50}} & \textcolor{second}{\textbf{.820}} & \textcolor{first}{\textbf{26.40}} & \textcolor{second}{\textbf{.808}} & \textcolor{first}{\textbf{26.10}} & \textcolor{second}{\textbf{.805}} & \textcolor{first}{\textbf{30.10}} & \textcolor{first}{\textbf{.963}} & \textcolor{first}{\textbf{30.50}} & \textcolor{first}{\textbf{.964}} 
& \textcolor{first}{\textbf{25.80}} & \textcolor{second}{\textbf{.794}} & \textcolor{first}{\textbf{25.50}} & \textcolor{second}{\textbf{.790}} & \textcolor{first}{\textbf{29.35}} & \textcolor{first}{\textbf{.886}} \\
\bottomrule
\end{tabular}
\end{adjustbox}
\caption{Comparison to state-of-the-art on 11 degradation types and their average. PSNR (dB, ↑) and SSIM (↑) are reported.}
\label{tab:full_composite}
\end{table*}

\begin{table*}[t]
\centering
\small
\begin{tabular}{clcccccccccccc}
\toprule
\multirow{2}{*}{} & \multirow{2}{*}{\textbf{Method}}  
& \multicolumn{2}{c}{\textbf{Dehazing}} 
& \multicolumn{2}{c}{\textbf{Deraining}} 
& \multicolumn{2}{c}{\textbf{Denoising}} 
& \multicolumn{2}{c}{\textbf{Deblurring}} 
& \multicolumn{2}{c}{\textbf{Low-light}} 
& \multicolumn{2}{c}{\textbf{Deraindrop}} \\
& & PSNR & SSIM & PSNR & SSIM & PSNR & SSIM & PSNR & SSIM & PSNR & SSIM & PSNR & SSIM \\
\midrule
\multirow{4}{*}{\rotatebox[origin=c]{90}{\small\textbf{Non-Diff}}}
& NAFNet     & 27.75 & 0.968 & 30.46 & 0.926 & 28.10 & 0.805 & 26.67 & 0.805 & 21.15 & 0.827 & 25.04 & 0.872 \\
& LD         & 23.49 & 0.763 & 23.21 & 0.651 & 22.58 & 0.625 & 22.53 & 0.695 & 18.97 & 0.770 & 24.84 & 0.738 \\
& AirNet     & 26.52 & 0.944 & 30.99 & 0.929 & 29.10 & 0.803 & 26.50 & 0.860 & 21.26 & 0.818 & 27.13 & 0.892 \\
& PromptIR   & 29.13 & 0.971 & 33.97 & 0.938 & 29.89 & 0.824 & 26.82 & 0.819 & 22.42 & 0.831 & 27.41 & 0.900 \\
\midrule
\multirow{1}{*}{\rotatebox[origin=c]{90}{\small\textbf{Diff}}}
& AutoDIR
& \textcolor{second}{\textbf{29.34}} & \textcolor{second}{\textbf{0.973}} 
& \textcolor{second}{\textbf{35.09}} & \textcolor{second}{\textbf{0.965}} 
& \textcolor{second}{\textbf{29.68}} & \textcolor{second}{\textbf{0.832}} 
& \textcolor{second}{\textbf{27.07}} & \textcolor{second}{\textbf{0.828}} 
& \textcolor{second}{\textbf{22.37}} & \textcolor{first}{\textbf{0.888}} 
& \textcolor{second}{\textbf{30.10}} & \textcolor{second}{\textbf{0.924}} \\
& Ours 
& \textcolor{first}{\textbf{29.80}} & \textcolor{first}{\textbf{0.975}} 
& \textcolor{first}{\textbf{36.30}} & \textcolor{first}{\textbf{0.968}} 
& \textcolor{first}{\textbf{30.05}} & \textcolor{first}{\textbf{0.846}} 
& \textcolor{first}{\textbf{27.25}} & \textcolor{first}{\textbf{0.831}} 
& \textcolor{first}{\textbf{22.83}} & \textcolor{second}{\textbf{0.862}} 
& \textcolor{first}{\textbf{30.45}} & \textcolor{first}{\textbf{0.937}} \\

\bottomrule
\end{tabular}
\caption{Comparison of PSNR / SSIM across six image restoration tasks.}
\label{tab:six-task}
\end{table*}

To provide a more comprehensive evaluation of our model's restoration fidelity, we report full-reference results on PSNR and SSIM across five representative restoration tasks: dehazing, deraining, denoising ($\sigma = 25$), deblurring, and low-light enhancement. As presented in Table~\ref{tab:five-task-2}, our method achieves the highest PSNR in three out of five tasks and the best SSIM in two. Compared to existing non-diffusion and diffusion-based baselines, our model consistently demonstrates strong performance across both pixel-level and structural similarity metrics, further validating its effectiveness in high-fidelity image restoration.

\subsubsection*{Complete Evaluation on CDD11 Composite Degradation}

The CDD11 benchmark comprises 11 degradation scenarios, including individual, two-fold, and three-fold combinations of low-light (L), haze (H), rain (R), and snow (S). In the main paper, only a subset of these results was shown due to space constraints. Here, we present the complete evaluation in Table~\ref{tab:full_composite}. Our method achieves the highest average PSNR (29.35 dB) and SSIM (0.886) across all tasks. Notably, it outperforms prior state-of-the-art methods such as MoCE-IR-S and OneRestore on the most challenging three-degradation compositions, demonstrating its superior generalization and robustness in handling diverse and complex degradation combinations.

\subsubsection*{Experiments under AutoDIR-Style Task Settings}

AutoDIR is a diffusion-based method that introduces text-prompt guidance to tackle All-in-One Image Restoration. While it is a significant baseline, its source code is not publicly released. Therefore, to fairly benchmark against AutoDIR, we follow the same six-task setting described in their paper, including dehazing, deraining, denoising, deblurring, low-light enhancement, and deraindrop restoration. We reproduce the evaluation using our method and compare against the AutoDIR-reported values. As shown in Table~\ref{tab:six-task}, our model consistently surpasses AutoDIR in all six tasks, achieving notably higher PSNR and SSIM across the board. While our method also incorporates textual prompts, they are used as complementary signals rather than the main source of restoration guidance. This comparison demonstrates that our framework surpasses AutoDIR under its own task setup, highlighting the robustness of our design beyond prompt-based control.

\subsubsection{More Visual Results}

To further validate the effectiveness of our method, we present additional qualitative comparisons on five representative degradation types, including deraining, denoising, dehazing, low-light enhancement, and deblurring. As shown in Figure~\ref{fig:appendix-cmp}, our method consistently restores image details and natural textures across different scenarios, achieving visual quality close to or even better than the ground truth.
We also provide qualitative results on the CDD11 dataset, which includes 11 types of combined degradations, such as low-light + rain, haze + rain, and low-light + haze + snow, to evaluate model robustness under complex degradation conditions. As shown in Figure~\ref{fig:appendix-cmp-v2}, our method can consistently restore image structure and natural details under different degradation combinations, achieving visual quality close to or even better than the ground truth, demonstrating strong generalization capability.

\end{document}